\documentclass[runningheads]{llncs}
\usepackage[T1]{fontenc}
\usepackage{amsmath}
\usepackage[breaklinks,colorlinks]{hyperref}
\hypersetup{colorlinks=true}
\usepackage{graphicx,verbatim}
\newcommand{\keypoint}[1]{\vspace{0.1cm}\noindent\textbf{#1}\quad}
\usepackage{colortbl}
\usepackage{subcaption}
\usepackage{color}
\definecolor{LightCyan}{rgb}{0.95,0.95,0.95}
\definecolor{LightGrey}{rgb}{0.85,0.91,0.98}
\usepackage{multirow}
\usepackage{booktabs}
\usepackage{enumitem}

\begin{document}
\title{Zero-shot Domain Generalization of Foundational Models for 3D Medical Image Segmentation:\\An Experimental Study}
\titlerunning{Zero-shot Domain Generalization for 3D Medical Segmentation FMs}

\author{Soumitri Chattopadhyay$^1$ \and Ba\c{s}ar Demir$^1$ \and Marc Niethammer$^2$}  
\authorrunning{Chattopadhyay et al.}
\institute{University of North Carolina at Chapel Hill \and University of California, San Diego \\
    \email{\{soumitri,bdemir\}@cs.unc.edu} \\
    \email{mniethammer@ucsd.edu}}

\maketitle              
\begin{abstract}
Domain shift, caused by variations in imaging modalities and acquisition protocols, limits model generalization in medical image segmentation. While foundation models (FMs) trained on diverse large-scale data hold promise for zero-shot generalization, their application to volumetric medical data remains underexplored. In this study, we examine their ability  towards domain generalization (DG), by conducting a comprehensive experimental study encompassing 6 medical segmentation FMs and 12 public datasets spanning multiple modalities and anatomies. Our findings reveal the potential of promptable FMs in bridging the domain gap via smart prompting techniques. Additionally, by probing into multiple facets of zero-shot DG, we offer valuable insights into the viability of FMs for DG and identify promising avenues for future research.


\end{abstract}
\section{Introduction}

Medical imaging involves a broad variety of scanners, acquisition protocols, imaging modalities (e.g., CT, MR, PET, ultrasound), sequences (T1w, T2w, ADC MRIs), and varying demographics. As a result, there are large variations in the distributions of the visual data. This issue is commonly referred to as \textit{domain shift}~\cite{domainshiftmri}, and poses a significant challenge for classical data-driven models, where the test distribution is assumed to match the training distribution. To overcome this, prior works have focused on domain adaptation~\cite{li2018adaptive,cyclegan3d}, test-time adaptation~\cite{valanarasu2024fly,karani2021test} and domain generalization (DG)~\cite{DG_MEDIA}. However, using these approaches requires individual models to be trained for a given pair of modalities and hence, cannot be adopted for generalized applications.



Originally rooted in the NLP community~\cite{fms_discuss,gpt2018,nlp_prompting}, foundational models (FMs) have recently been developed in the visual domain~\cite{fm_vison_tpami}, for vision-language modeling~\cite{clip}, visual generation~\cite{dalle}, registration~\cite{unigradicon,mmgradicon} and segmentation~\cite{sam,medsam}. A key characteristic of these models is that they are trained on large-scale curated data, which suggests \textit{the potential to generalize to new distributions in a zero-shot manner}. In particular, CLIP~\cite{clip} and SAM~\cite{sam} have shown impressive zero-shot capabilities for coarse-level classification and sparse-promptable segmentation tasks~\cite{sam_survey}.  Recently, the SAM paradigm has been extended to medical volumes, leading to large-scale volumetric segmentation models~\cite{sammed3d,segvol}, going beyond 2D slice-level approaches~\cite{medsam} to full-volume inference. However, to our best knowledge, no studies yet have investigated these foundational 3D segmenters for \textit{zero-shot domain generalization}.

In this work, we probe the feasibility of zero-shot domain generalization for 3D medical image segmentation with segmentation FMs. To this end, we first describe a taxonomy that categorizes existing off-the-shelf FMs, based on their training knowledge (specific modalities/general), inputs (promptable/automatic), and targets (finite set of classes, or prompt-specific). Our study encompasses 6 medical FMs and 12 volumetric segmentation datasets (described in Sec.~\ref{sec:materials}) comprising various anatomies and modalities. To assess zero-shot DG holistically, we examine dataset distribution shifts, cross-modal transferability (Sec.~\ref{sec:domainfms}), and generalization to unseen anatomies (Sec.~\ref{sec:visprompt}). Our empirical findings reveal that most existing FMs exhibit a significant domain gap compared to in-domain-trained specialist models~\cite{nnunet}. However, recent text-promptable models show stronger DG, narrowing this gap (Sec.~\ref{sec:textprompt}), potentially making them a promising direction for domain-generalizable models in medical image computing.


Our study is driven by two key factors: (i) Zero-shot generalizability is a desirable trait of FMs~\cite{fm_vison_tpami}, and given their success in natural vision tasks~\cite{clip,sam,sam_survey}, it is crucial to investigate whether similar performance extends to medical image segmentation across datasets, anatomies, and modalities. (ii) DG in medical image segmentation remains a long-standing challenge with significant relevance to federated learning~\cite{liu2021feddg,nguyen2022fedsr}, automating the manual clinical segmentation process, and adoption to newer modalities with limited annotations. Our work is both timely and essential for advancing research on zero-shot FM generalization.



\vspace{0.1cm}
\noindent \textbf{In summary, the contributions of our work are as follows:} 
\vspace{-0.08cm}

\begin{enumerate}
    \item We define a taxonomy for existing 3D medical segmentation FMs structured around domain generalization.
    \item We assess their zero-shot performance across datasets encompassing diverse anatomies and multiple modalities/sequences.  
    \item Our analysis covers multiple facets of DG, including generalization to unseen modalities, the impact of different prompts, and adaptation to novel anatomical structures.
    \item Our findings provide key insights into the feasibility of FMs for domain generalization and highlight promising directions for future research.
\end{enumerate}

\section{Related Work}

\keypoint{Foundational models for segmentation:} While prior approaches relied on training segmentation models based on a finite set of target classes~\cite{nnunet,fseft,mrsegmentator} which limits their wider applicability, the advent of large-scale promptable models like SAM~\cite{sam} imparted flexibility by enabling the model to generate outputs based on user-specified inputs. Recently, this paradigm has been adopted in medical imaging for 2D~\cite{medsam,sam_med2d} and 3D~\cite{sammed3d,segvol} segmentations, being trained on assembled corpora of multi-sourced datasets. \emph{It is useful to probe how well, if at all, these models can generalize to varying data distributions and imaging modalities -- which is what we explore in our current study.}

\keypoint{Domain generalization in medical imaging:} The broad range of imaging modalities, scanners, protocols, and demographics make domain shift a prevalent issue for 3D medical image segmentation~\cite{visionfmseccv}, with dataset-specific trained models struggling to generalize to different domains. Prior approaches have focused on unsupervised domain adaptation~\cite{cyclegan3d,li2018adaptive}, test-time adaptation~\cite{valanarasu2024fly,karani2021test} and domain generalization~\cite{DG_MEDIA,dou2019domain,fu2024cosam}. The recent advent of FMs for medical imaging~\cite{fseft,vista3d,sammed3d} makes them potential candidates to achieve domain generalization, owing to their large-scale training across diverse datasets. \emph{In this work, we investigate the generalizability of such models for 3D segmentation tasks across multiple domain shift scenarios.}

\section{Materials and Methods}\label{sec:materials}


\subsection{Datasets}

We use several publicly available volumetric medical segmentation datasets spanning multiple anatomical regions and modalities, which facilitate (1) \underline{zero-shot} \underline{domain transfer} and (2) \underline{unseen organ segmentation} experiments, both of which we feel are crucial to probe domain generalization.

\begin{enumerate}[label=\textbullet]
    \item \textbf{Abdominal organs:} Datasets containing segmentations of various organs in the human abdominal region -- BTCV~\cite{btcv}, AMOS`22~\cite{amos22}, FLARE`22~\cite{flare22}, CHAOS~\cite{chaos}, MSD-Spleen~\cite{decathlon}. Notably, AMOS`22 provides both CT and MR datasets having the \textit{same target organs}, and we use both in our study.
    
    \item \textbf{Pelvic anatomy:} Focused on prostate segmentation from MR volumes -- MSD-Prostate~\cite{decathlon} (having T2w and ADC sequences) and PROMISE12~\cite{promise12}. We merged the peripheral and transitional regions in MSD-Prostate into a single region for consistency with PROMISE12 and models trained on the full prostate.
    
    \item \textbf{Cardiac substructures:} Datasets with segmentations of the atria, ventricles, and myocardium -- MSD-Heart~\cite{decathlon} and MMWHS - CT \& MR~\cite{mmwhs}.
    
\end{enumerate}

\noindent \autoref{tab:datasets} provides more details on the composition of the datasets. 

\begin{table}[h]
    \centering
    \caption{Datasets used in our domain generalization study. They encompass a diverse range of anatomies (abdomen, pelvic, cardiac and entire body) across multiple modalities (CT, MR) and sequences (e.g., T1w, T2w, T2-SPIR, and ADC MRIs).} 
    \label{tab:datasets}
    \setlength{\tabcolsep}{5pt}
    \renewcommand{\arraystretch}{1.2}
    \resizebox{\columnwidth}{!}{
    \begin{tabular}{ccccc}
         \toprule
         \textbf{Dataset} & \textbf{Anatomies} & \textbf{Modalities} & \textbf{\#classes} & \textbf{\#samples} \\
         \midrule
         BTCV \cite{btcv} & abdominal organs & CT & 13 & 30 \\
         AMOS`22 \cite{amos22} & abdominal organs & CT, MR & 15 & CT: 120, MR: 60 (test set)  \\
         FLARE`22 \cite{flare22} & abdominal organs & CT & 13 & train: 50; test: 22 \\
         CHAOS \cite{chaos} & abdominal organs & MR: T2-SPIR & 4 & 20 \\
         MSD-Spleen \cite{decathlon} & spleen (abdomen) & CT & 1 & 41 \\
         MSD-Prostate \cite{decathlon} & prostate (pelvic) & MR: T2w, ADC & 1 & 32 \\
         PROMISE12 \cite{promise12} & prostate (pelvic) & MR: T2w & 1 & train: 50; test: 30 \\
         MMWHS \cite{mmwhs} & cardiac substructures & CT, MR & 5 & CT: 20, MR: 20 \\
         MSD-Heart \cite{decathlon} & left atrium (cardiac) & MR & 1 & 20 \\
         \bottomrule
    \end{tabular}}
\end{table}

\subsection{Model Taxonomy}

We describe a taxonomy for the 3D segmentation FMs used in this study and list them below. Additionally, \autoref{tab:models} describes their training domains, target classes and preprocessing of inputs before passing them into their networks.

\keypoint{Domain-specific FMs:} These include models that have been trained on curated corpora comprising multiple anatomy-specific datasets in a specific modality. In this category, we consider the following models: FSEFT~\cite{fseft}, VISTA3D~\cite{vista3d} and MRSegmentator~\cite{mrsegmentator}. While the former two models were trained with publicly accessible CT volumes only, MRSegmentator was trained with both CT and MR data, the latter coming from an in-house corpus~\cite{mrsegmentator}. Moreover, FSEFT focuses on segmenting abdominal and thoracic structures, whereas VISTA3D and MRSegmentator  are trained on a larger set of organs.  

\keypoint{Visual Prompted FMs:} As discussed earlier, promptable segmentation models~\cite{sam,sammed3d} enable more general segmentations,  since the target can be guided by user-defined visual region of interest indicators (e.g., points/clicks, bounding boxes, or segmentation masks)~\cite{sam}. Following this paradigm, we evaluate two such recently proposed models for volumetric medical segmentation: SAM-Med3D~\cite{sammed3d} and SegVol~\cite{segvol}. While SAM-Med3D is trained with 3D point prompts and multiple modalities, the latter is trained only on CT volumes and enables both points and bounding boxes as visual prompt inputs.

\keypoint{Text Prompted FMs:} Moving beyond spatial prompts, we also evaluate text promptable segmentation models -- SegVol~\cite{segvol} and SAT~\cite{sat}, where spatial prompts can be replaced (or augmented) with semantic text for target organs.

\begin{table}[t]
    \centering
    \caption{List of foundational 3D medical segmentation models, along with their respective input preprocessing. We focus on domain-specific trained FMs, visual and text promptable FMs and investigate their domain generalization feasibility.} 
    \label{tab:models}
    \setlength{\tabcolsep}{4pt}
    \renewcommand{\arraystretch}{1.2}
    \resizebox{\columnwidth}{!}{
    \begin{tabular}{ccccc}
         \toprule
         \textbf{Model} & \textbf{Training domain} & \textbf{\#classes} & \textbf{Resampling} & \textbf{Intensity clipping ranges} \\
         \midrule
         FSEFT \cite{fseft} & CT & 29 &  $1.5mm^3$ & CT: $[-175,250]$ HU; MR: $[1,99]^{th}$ \%ile  \\
         VISTA3D \cite{vista3d} & CT & 124 & $1.5mm^3$ & CT: $[-1000,1000]$ HU; MR: $[1,99]^{th}$ \%ile  \\
         MRSegmentator \cite{mrsegmentator} & CT, MR & 40 & $1.5\times1.5\times3.0mm^3$ & CT: $[0.5,99.5]^{th}$ \%ile; MR: Z-score norm. \\
         SAM-Med3D \cite{sammed3d} & CT, MR & \textbf{--} & $1.5mm^3$ & CT \& MR: Z-score norm. \\
         SegVol \cite{segvol} & CT & \textbf{--} & \textbf{--} & CT: $[0.5,99.5]^{th}$ \%ile; MR: $[1,99]^{th}$ \%ile \\
         SAT \cite{sat} & CT, MR, PET & \textbf{--} & $1.0\times1.0\times3.0mm^3$ & CT: $[-500,1000]$ HU; MR: $[0.5,99.5]^{th}$ \%ile \\
         \bottomrule
    \end{tabular}}
\end{table}


\section{Experiments}\label{expts}

\keypoint{Implementation:}All models are implemented in PyTorch~\cite{pytorch} and MONAI~\cite{monai}, accelerated by a 48GB Nvidia RTX A6000 GPU. We used publicly available codebases and checkpoints for all models. Specific preprocessing details for each model, such as input spacing and intensity ranges for different modalities, are outlined in \autoref{tab:models}. \textit{Our inference scripts will be open-sourced upon acceptance.}


\keypoint{Evaluation metric:}We used the Dice Similarity Coefficient (DSC)~\cite{dice} to report all experimental results, considering it to be the de facto standard metric for evaluating segmentation models~\cite{visionfmseccv,nnunet_revisited}.

\keypoint{In-domain specialist model:}We chose nnUNet \cite{nnunet} as the specialist ``oracle'' model defining the empirical \textit{upper bound} performance in-domain, subsequently quantifying the \textit{domain gap} with respect to the FMs. Choosing nnUNet is supported by a recent study~\cite{nnunet_revisited} that concluded it to be the SoTA segmentation model. \textit{Specifically, we report results from SAT}~\cite{sat}, \textit{which extensively trained dataset-specific nnUNets and provided full training details for reproducibility.}

\subsection{Zero-shot transfer of domain-specific FMs}\label{sec:domainfms}

Our analysis starts with probing the domain transferability of large-scale modality-specific trained FMs: FSEFT~\cite{fseft}, VISTA3D~\cite{vista3d} and MRSegmentator~\cite{mrsegmentator}. In \autoref{tab:sec4p1}, we report average Dice scores across all organs for each dataset.  

\begin{table}[h]
    \centering
    \caption{Performance of domain-specific FMs across different datasets. Datasets marked as (\textbf{--}) were used for training the respective model and are hence excluded for domain generalization experiments. Cardiac and Prostate datasets are excluded since those anatomies were not present during training of the comparative models.} 
    \label{tab:sec4p1}
    \setlength{\tabcolsep}{5pt}
    \resizebox{\columnwidth}{!}{
    \begin{tabular}{lcccccc}
        \toprule
        \multirow{2}{*}{{Domain-specific FMs}} & \multicolumn{4}{c}{\textbf{CT}} & \multicolumn{2}{c}{\textbf{MR}} \\ 
        \cmidrule(lr){2-5} \cmidrule(lr){6-7} 
        & {BTCV} & {AMOS`22} & {FLARE`22} & {MSD-Spleen} & {CHAOS} & {AMOS`22} \\
        \midrule
        FSEFT \cite{fseft} & \textbf{--} & \textbf{--} & 75.06 & \textbf{--} & 23.76 & 38.72 \\
        VISTA3D \cite{vista3d} & 82.52 & \textbf{--} & 87.29 & \textbf{--} & 24.10 & 42.28 \\
        MRSegmentator \cite{mrsegmentator} & \textbf{84.42} & \textbf{84.35} & \textbf{88.85} & \textbf{95.68} & \textbf{89.83} & \textbf{74.86} \\
        \midrule
        \rowcolor{LightCyan}
        nnUNet \textbf{(specialist)} \cite{nnunet} & 88.89 & 89.77 & 93.36 & 96.70 & 88.89 & 86.43 \\
        \bottomrule
    \end{tabular}}
\end{table}

\keypoint{Analysis.}Notably, we observe significant domain gaps for all these models compared to the in-domain specialists~\cite{nnunet}, for both modality transfer (CT$\xrightarrow{}$MR) as well as distribution shifts within the same modality. Expectedly, the CT-trained FSEFT and VISTA3D models greatly struggle to generalize to MR datasets having on same target anatomies they were trained on, and yield low segmentation scores on CHAOS-MR ($\approx65\%$ lower Dice) and AMOS-MR ($\approx46\%$ lower Dice) compared to the in-domain nnUNet. Furthermore, FSEFT also fails to generalize within CT, performing $\approx18\%$ lower on FLARE’22 than nnUNet, while VISTA3D performs better but still lags by $\approx6\%$ behind on BTCV and FLARE’22. MRSegmentator shows superior generalization, significantly reducing the performance gap with in-domain specialists. However, since it was trained on both CT and MR~\cite{mrsegmentator}, this improvement is expected. \emph{The key takeaway is that while FSEFT and VISTA3D struggle for both within-modality and unseen-modality distribution shifts, MRSegmentator improves domain generalization with stronger performance on within-modality distribution drifts. Nevertheless, for most datasets, even MRSegmentator shows a considerable segmentation performance gap compared to the dataset-specific nnUNets.}



\subsection{Visual promptable FMs}\label{sec:visprompt}

With domain-specific FMs having limited target organs, we next shift focus to what has been the recent paradigm of segmentation FMs -- prompting with spatial region of interest  inputs~\cite{sam,sammed3d,segvol}. Our candidate models are SAM-Med3D~\cite{sammed3d} and SegVol~\cite{segvol}, with 3D points and bounding boxes as prompts. \autoref{tab:sec4p2} shows the average Dice scores for each testing dataset.

\begin{table}[h]
    \centering
    \caption{Performance of visually promptable foundational models. `P' denotes the number of points provided as prompts. Values are Dice \%.} 
    \label{tab:sec4p2}
    \setlength{\tabcolsep}{7pt}
    \resizebox{\columnwidth}{!}{
    \begin{tabular}{lccccccc}
        \toprule
        \multirow{2}{*}{Visual Prompted FMs} & \multicolumn{4}{c}{{SAM-Med3D} \cite{sammed3d}} & \multicolumn{2}{c}{{SegVol} \cite{segvol}} & \multirow{2}{*}{{nnUNet} \cite{nnunet}} \\
        \cmidrule(lr){2-5} \cmidrule(lr){6-7}
        & P=1 & P=3 & P=5 & P=10 & P=5 & Bbox & {\cellcolor{LightCyan}\textbf{(specialist)}} \\
        \midrule
        \multicolumn{8}{l}{\textbf{Abdominal CT}} \\
        BTCV & 78.99 & 80.99 & 81.49 & \textbf{81.86} & 67.04 & 80.15 & {\cellcolor{LightCyan}88.89} \\
        AMOS`22 & 79.06 & 81.74 & 82.31 & \textbf{83.18} & 65.46 & 79.82 & {\cellcolor{LightCyan}89.77} \\
        FLARE`22 & 84.41 & 86.63 & 86.91 & \textbf{87.28} & 68.25 & 76.93 & {\cellcolor{LightCyan}93.36} \\
        MSD-Spleen & 94.00 & 94.43 & 94.51 & \textbf{94.62} & 91.72 & 94.21 & {\cellcolor{LightCyan}96.70} \\
        \midrule
        \multicolumn{8}{l}{\textbf{Abdominal MR}} \\
        CHAOS & 89.52 & 90.64 & 91.54 & \textbf{91.76} & 80.25 & 10.11 & {\cellcolor{LightCyan}88.89} \\
        AMOS`22 & 71.01 & 75.04 & 76.43 & \textbf{77.47} & 60.18 & 45.96 & {\cellcolor{LightCyan}86.43} \\
        \midrule
        \multicolumn{8}{l}{\textbf{Pelvic (MR)}} \\
        MSD-Prostate (T2) & 84.32 & 88.33 & 88.85 & \textbf{89.93} & 57.53 & 37.27 & {\cellcolor{LightCyan}87.60} \\
        MSD-Prostate (ADC) & 80.53 & 84.77 & 86.37 & \textbf{86.97} & 66.76 & 25.82 & {\cellcolor{LightCyan}87.60} \\
        PROMISE12 & 84.41 & 86.86 & 88.02 & \textbf{88.98} & 55.27 & 39.42 & {\cellcolor{LightCyan}88.86} \\
        \midrule
        \multicolumn{8}{l}{\textbf{Cardiac}} \\
        MMWHS (CT) & 52.09 & 62.25 & 65.24 & \textbf{69.18} & 46.82 & 61.53 & {\cellcolor{LightCyan}88.64} \\
        MMWHS (MR) & 54.55 & 66.67 & 70.08 & \textbf{72.99} & 39.82 & 51.48 & {\cellcolor{LightCyan}30.88} \\
        MSD-Heart (MR) & 79.69 & 86.32 & 87.20 & \textbf{88.22} & 42.66 & 57.88 & {\cellcolor{LightCyan}94.28} \\
        \bottomrule
    \end{tabular}}
\end{table}

\keypoint{Effect of prompts.} Unsurprisingly, for SAM-Med3D, segmentation performance consistently improves with an increase in point prompts. However, it is interesting to observe that the performance gain quickly saturates for even greater increase in number of points: for instance, for cardiac substructures~\cite{mmwhs,decathlon}, \autoref{tab:sec4p2} shows that while increasing $P=1\xrightarrow{}3$ greatly improves segmentation performance ($\Delta\approx10\%$) -- the margin of improvement is lower for $P=3\xrightarrow{}5$ ($\Delta\approx2.6\%$) and $P=5\xrightarrow{}10$ ($\Delta\approx2.3\%$). Other datasets show lower improvement margins with an increasing number of points. For SegVol, point-prompted scores remain consistently lower than for SAM-Med3D, but bounding boxes improve segmentation on several datasets. Notably, for all CT datasets, bounding boxes outperform points, aligning with prior findings~\cite{segvol}. However, for unseen MR modalities, points outperform boxes (e.g., AMOS`22-MR: 60.18 vs. 45.96, PROMISE12: 55.27 vs. 39.42). We hypothesize that bounding boxes capture background regions, leading to misclassification of foreground voxels in unseen modalities, whereas points—always marking the foreground—avoid this issue.




\keypoint{Domain generalization analysis.} SegVol is a CT-only trained model; for zero-shot transfer on MR datasets, it performs considerably worse compared to SAM-Med3D with the same number of point prompts ($\Delta_{\text{P=5}}\text{(CHAOS)}\approx11\%, \Delta_{\text{P=5}}\text{(AMOS)}\approx16\%$). On the brighter side, SegVol achieves higher Dice scores than its domain-specific counterparts (FSEFT~\cite{fseft} and VISTA3D~\cite{vista3d}, from \autoref{tab:sec4p1}). SAM-Med3D was trained with both CT and MR data and shows stronger generalizability, even outperforming the specialist nnUNet on CHAOS (91.76 vs. 88.89) and MSD-Prostate T2 (89.93 vs. 87.60). Yet, it exhibits a significant domain gap against the in-domain specialist nnUNet on the abdominal ($\Delta_{\text{BTCV}}\approx7\%, \Delta_{\text{AMOS-MR}}\approx9\%)$ and cardiac ($\Delta_{\text{MMWHS-CT}}\approx20\%,\Delta_{\text{MSD-Heart}}\approx6\%$) datasets. Nevertheless, visual prompted FMs dominate domain-specific FMs in terms of \textit{applicability towards a broader range of anatomies}, paving the way to reduce the existing domain gap compared to specialist models. 

\keypoint{Generalization to \textit{unseen anatomies} via prompting.}SegVol’s training corpus lacked pelvic and cardiac anatomies~\cite{segvol}, allowing us to assess its zero-shot performance. From \autoref{tab:sec4p2}, for MMWHS-CT (known modality), SegVol with boxes nears SAM-Med3D’s performance with 3-point prompts (61.53 vs. 62.25), but SAM-Med3D surpasses it beyond this. For prostate and cardiac MR, scores fall well below in-domain specialist nnUNets~\cite{nnunet}. These results indicate that \textit{current 3D promptable FMs still struggle with unseen anatomies}, requiring further research and improvements.


\subsection{Text promptable FMs}\label{sec:textprompt}

Going further, recent works have also enabled semantic textual prompts to denote target structures for segmentation, typically processed via a pre-trained text encoder \cite{clip,bert,medcpt}. We have two candidate models -- SegVol with text prompts \cite{segvol} and SAT \cite{sat}. \autoref{tab:sec4p3} shows the corresponding Dice scores obtained.

\begin{table}[th]
    \centering
    \caption{Performance of text promptable FMs. SAT-Nano and SAT-Pro denote smaller (110M) and larger capacity (447M) model variants respectively, while SAT-Ft is a fine-tuned version \textit{(not zero-shot)} of SAT-Pro on the respective datasets.} 
    \label{tab:sec4p3}
    \setlength{\tabcolsep}{5pt}
    \resizebox{\columnwidth}{!}{
    \begin{tabular}{lccccccc}
        \toprule
        \multirow{2}{*}{Text Prompted FMs} & \multicolumn{3}{c}{{SegVol} \cite{segvol}} & \multicolumn{3}{c}{{SAT} \cite{sat}} & \multirow{2}{*}{{nnUNet} \cite{nnunet}} \\
        \cmidrule(lr){2-4} \cmidrule(lr){5-7}
        & text & text+bbox & text+points & Nano & Pro & {\cellcolor{LightGrey}Ft} & {{\cellcolor{LightCyan}}\textbf{(specialist)}} \\
        \midrule
        \multicolumn{8}{l}{\textbf{Abdominal CT}} \\
        BTCV & 80.68 & \textbf{81.87} & 71.70 & 79.60 & 80.71 & {\cellcolor{LightGrey}81.60} & {\cellcolor{LightCyan}88.89} \\
        AMOS`22 & 75.30 & 81.12 & 69.09 & 84.93 & \textbf{86.37} & {\cellcolor{LightGrey}88.75} & {\cellcolor{LightCyan}89.77} \\
        FLARE`22 & 77.32 & 78.73 & 72.79 & 88.79 & \textbf{91.12} & {\cellcolor{LightGrey}91.78} & {\cellcolor{LightCyan}93.36} \\
        MSD-Spleen & 95.75 & \textbf{95.79} & 95.76 & 93.50 & 94.12 & {\cellcolor{LightGrey}94.97} & {\cellcolor{LightCyan}96.70} \\
        \midrule
        \multicolumn{8}{l}{\textbf{Abdominal MR}} \\
        CHAOS & 73.23 & 21.32 & 81.06 & 82.07 & \textbf{87.28} & {\cellcolor{LightGrey}87.99} & {\cellcolor{LightCyan}88.89} \\
        AMOS`22 & 62.73 & 47.29 & 61.59 & 78.76 & \textbf{78.90} & {\cellcolor{LightGrey}84.82} & {\cellcolor{LightCyan}86.43} \\
        \midrule
        \multicolumn{8}{l}{\textbf{Pelvic (MR)}} \\
        MSD-Prostate (T2) & 0.01 & 20.53 & 26.65 & 73.38 & \textbf{78.33} & {\cellcolor{LightGrey}77.98} & {\cellcolor{LightCyan}87.60} \\
        MSD-Prostate (ADC) & 0.05 & 35.21 & 65.71 & 73.38 & \textbf{78.33} & {\cellcolor{LightGrey}77.98} & {\cellcolor{LightCyan}87.60} \\
        PROMISE12 & 0.20 & 24.68 & 30.50 & 84.55 & \textbf{86.51} & {\cellcolor{LightGrey}87.28} & {\cellcolor{LightCyan}88.86} \\
        \midrule
        \multicolumn{8}{l}{\textbf{Cardiac}} \\
        MMWHS (CT) & 0.03 & 35.54 & 13.01 & 88.23 & \textbf{89.97} & {\cellcolor{LightGrey}91.14} & {\cellcolor{LightCyan}88.64} \\
        MMWHS (MR) & 0.07 & 29.24 & 13.78 & 84.37 & \textbf{86.70} & {\cellcolor{LightGrey}87.73} & {\cellcolor{LightCyan}30.88} \\
        MSD-Heart (MR) & 0.02 & 54.45 & 37.16 & 90.28 & \textbf{92.61} & {\cellcolor{LightGrey}93.38} & {\cellcolor{LightCyan}94.28} \\
        \bottomrule
        \end{tabular}}
\end{table}

\begin{figure}[h]
    \centering
    \includegraphics[width=\columnwidth]{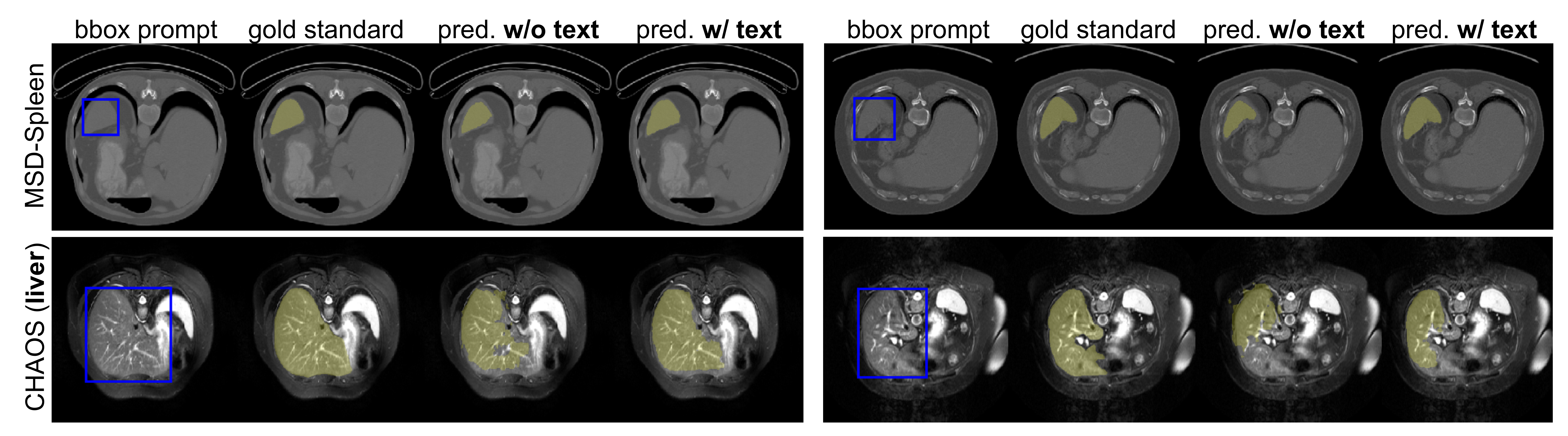}
    \caption{Qualitative results showing segmentation improvement in SegVol with \textbf{text + bbox} prompts for seen (CT; MSD-Spleen) and unseen (MR; CHAOS) modalities.}
    \label{fig:bbox_text}
\end{figure}

\keypoint{Text inputs boost spatial prompts.} We investigate the effect of semantic prompts as an additional guidance alongside spatial prompts for SegVol~\cite{segvol}, denoted by ``text+bbox'' and ``text+points'' in \autoref{tab:sec4p3}. Comparing them with corresponding vanilla spatial prompts in \autoref{tab:sec4p2}, we notice that for known anatomies (abdominal CT and MR), the addition of text consistently improves segmentations from raw spatial prompts (BTCV: 80.15 w/ bbox, 81.87 w/ text+bbox; CHAOS: 80.25 w/ points, 81.06 w/ text+points). We show example segmentation results in \autoref{fig:bbox_text}. For unseen anatomies, text prompts completely fail, since the model has no knowledge to spatially ground such concepts. 



\keypoint{Semantic prompts show better domain generalization.} As can be seen from \autoref{tab:sec4p3} (as well as \autoref{tab:sec4p2}), for known anatomies, text-prompted segmentation shows better generalizability across modalities, surpassing visual prompts. In particular, SAT shows strong segmentation performance across all datasets, reducing the empirical domain gap with in-domain specialists in several cases. SAT also outperforms SegVol in the text-prompted configuration on most datasets. However, we point out that SAT and SegVol are not directly comparable: the training corpus of SAT is significantly larger than for SegVol~\cite{sat}, and while SegVol uses a CLIP~\cite{clip} encoder for text, while SAT leverages a BERT~\cite{bert} model pre-trained on medical documents, enhancing anatomical knowledge during prompting. Although there still remains a considerable empirical gap of SAT with in-domain nnUNets, it has been able to reduce it further as compared to previous FMs~\cite{fseft,vista3d,sammed3d} with greater applicability across different anatomies and modalities, which makes us optimistic that text-promptable segmentation~\cite{sat,clip_driven_universal} possibly combined with visual prompts for further refinement will eventually become the universal segmentation paradigm.

\section{Conclusion}


In this study, we explored the DG feasibility of FMs for 3D medical image segmentation, encompassing a variety of domain shift cases across anatomies and modalities, and different categories of FMs for a holistic analysis. The take-home message from our empirical findings is -- while domain-specific FMs fail to generalize across domain shifts, promptable FMs show potential to bridge the domain gap, but require smart prompting techniques that can effectively convey the region-of-interest pointer to the model, possibly achievable by text prompts with spatial inputs for refinement. Future works should build upon these findings to develop a truly universal, domain-generalized segmentation framework.


\section{Acknowledgements}

This research was, in part, funded by the National Institutes of Health (NIH) under other transactions 1OT2OD038045-01 and NIAMS 1R01AR082684. The views and conclusions contained in this document are those of the authors and should not be interpreted as representing official policies, either expressed or implied, of the NIH.

%
%
%
\bibliographystyle{splncs04}
\bibliography{references}

\appendix
\section{Organ-wise segmentation results}

\begin{figure}
    \centering
    \makebox[\textwidth][c]{%
        \includegraphics[width=1.2\textwidth]{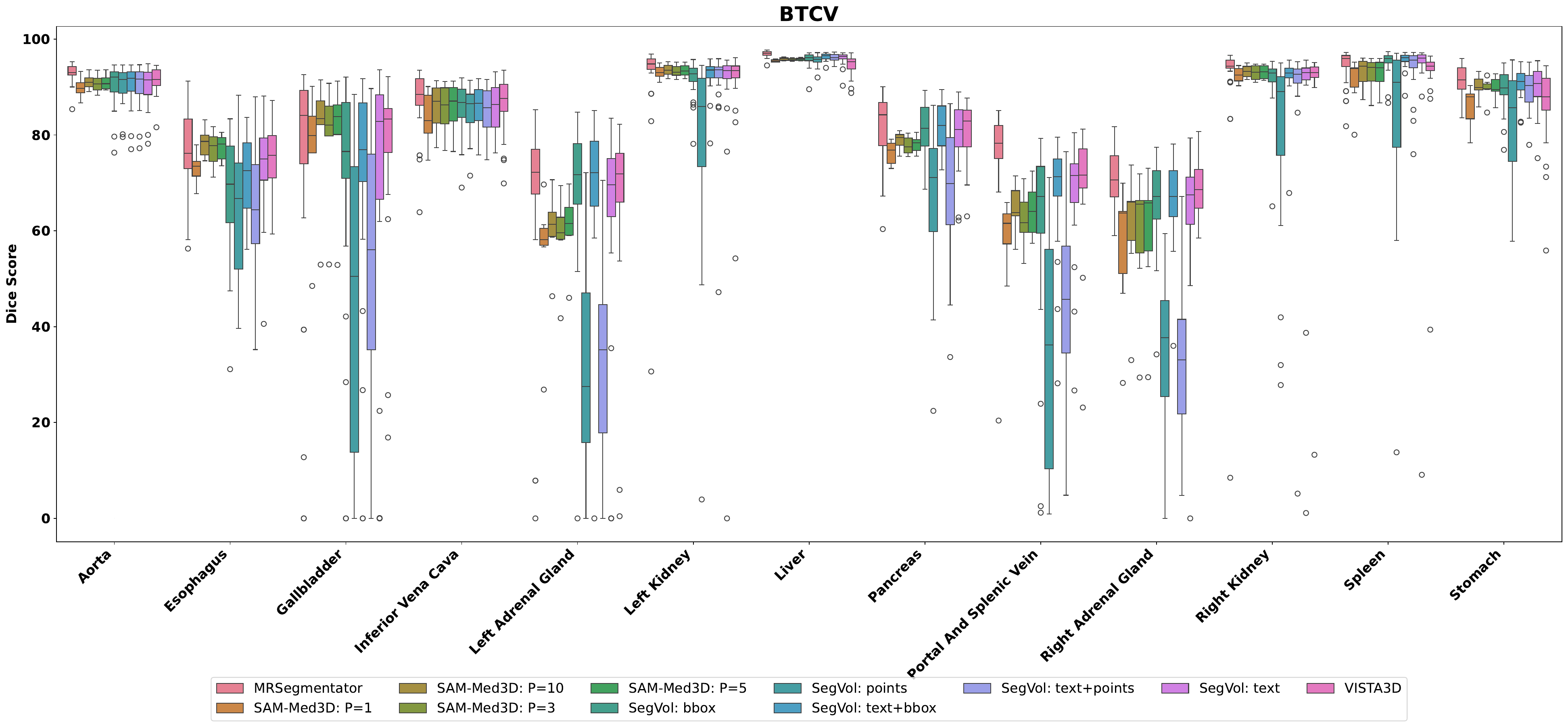}
    }
    \makebox[\textwidth][c]{%
        \includegraphics[width=1.2\textwidth]{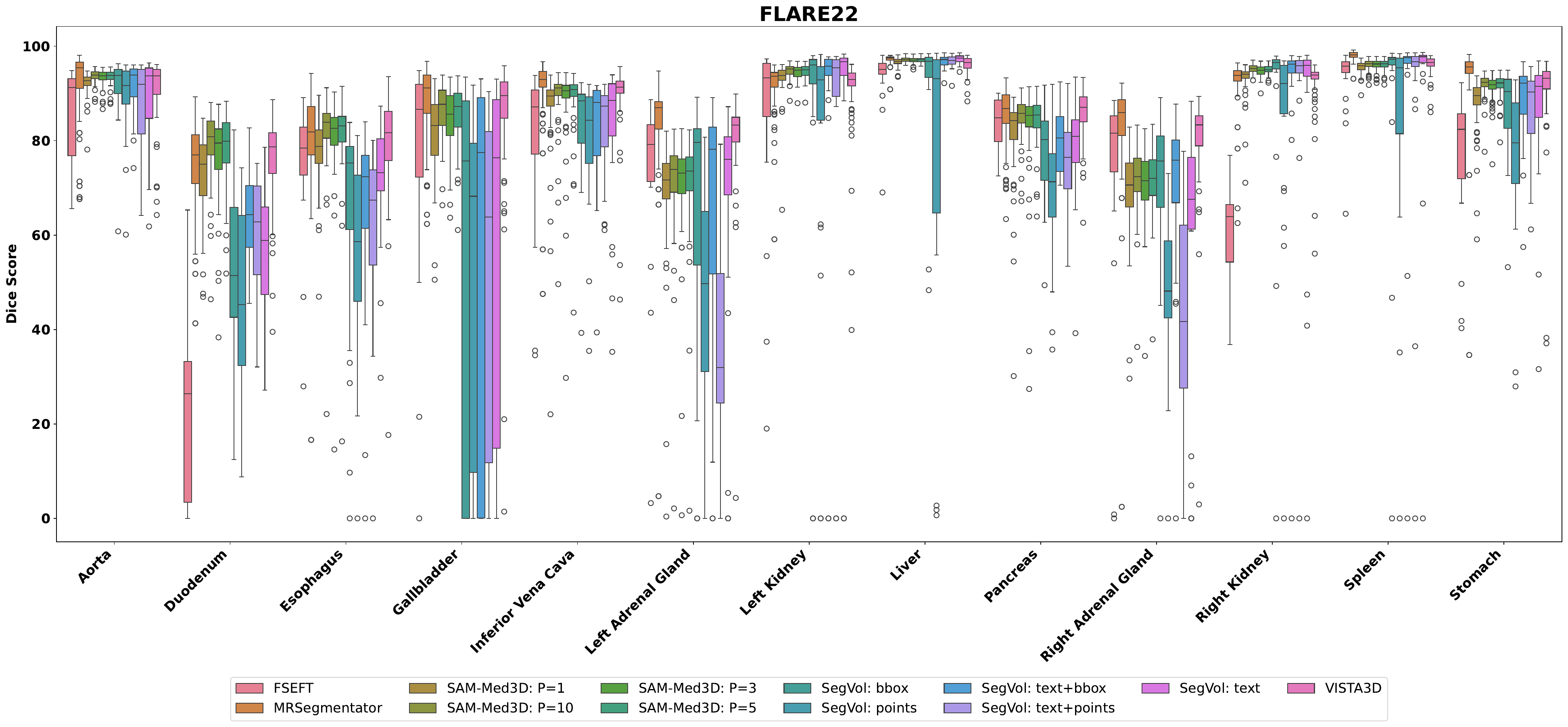}
    }
    \makebox[\textwidth][c]{%
        \includegraphics[width=1.2\textwidth]{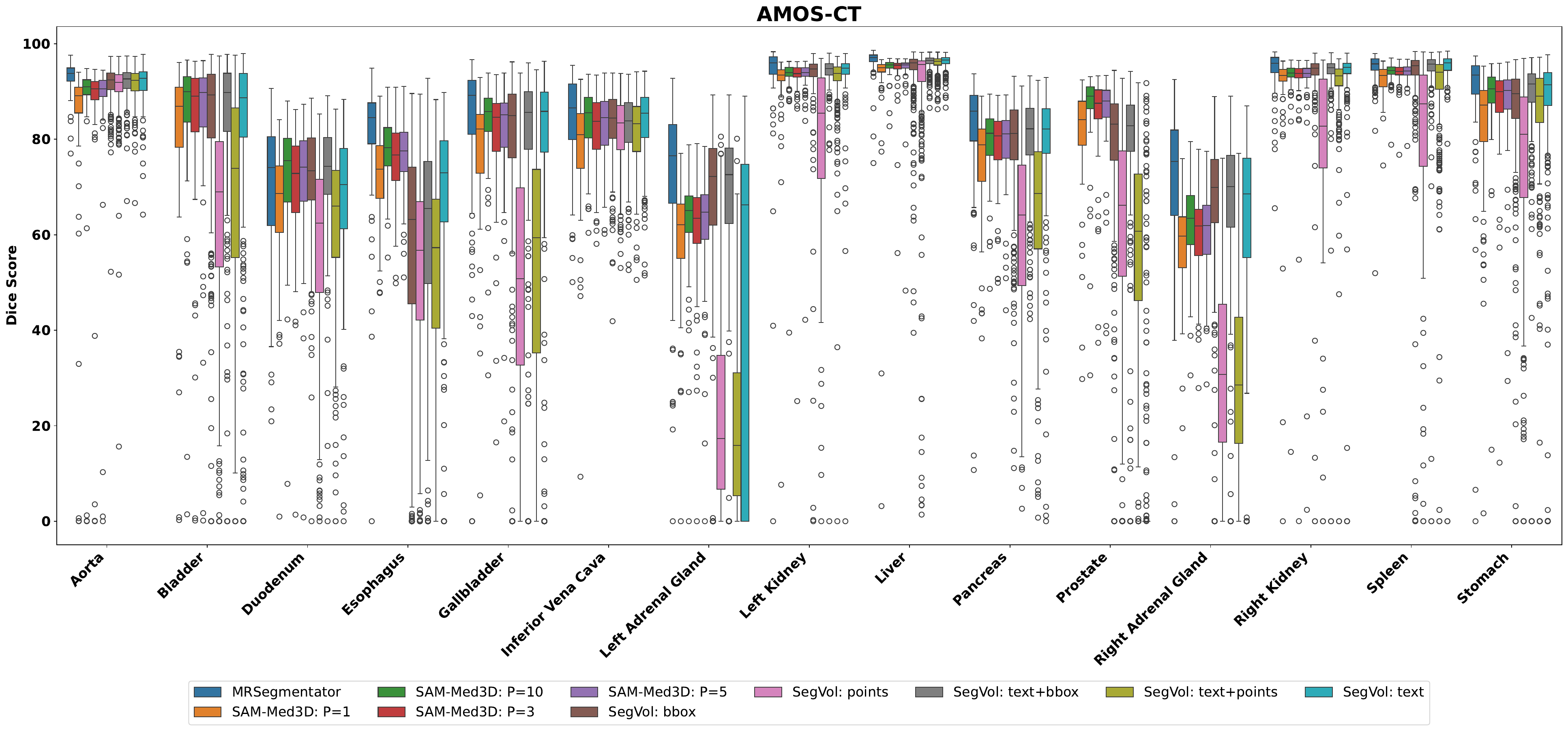}
    }
    \caption{Box plots for organ-wise dice scores for all comparative models on the \textbf{abdominal CT} datasets used in this study.}
    \label{fig:organwise_abdct}
\end{figure}

\begin{figure}
    \centering
    \makebox[\textwidth][c]{%
        \includegraphics[width=1.2\textwidth]{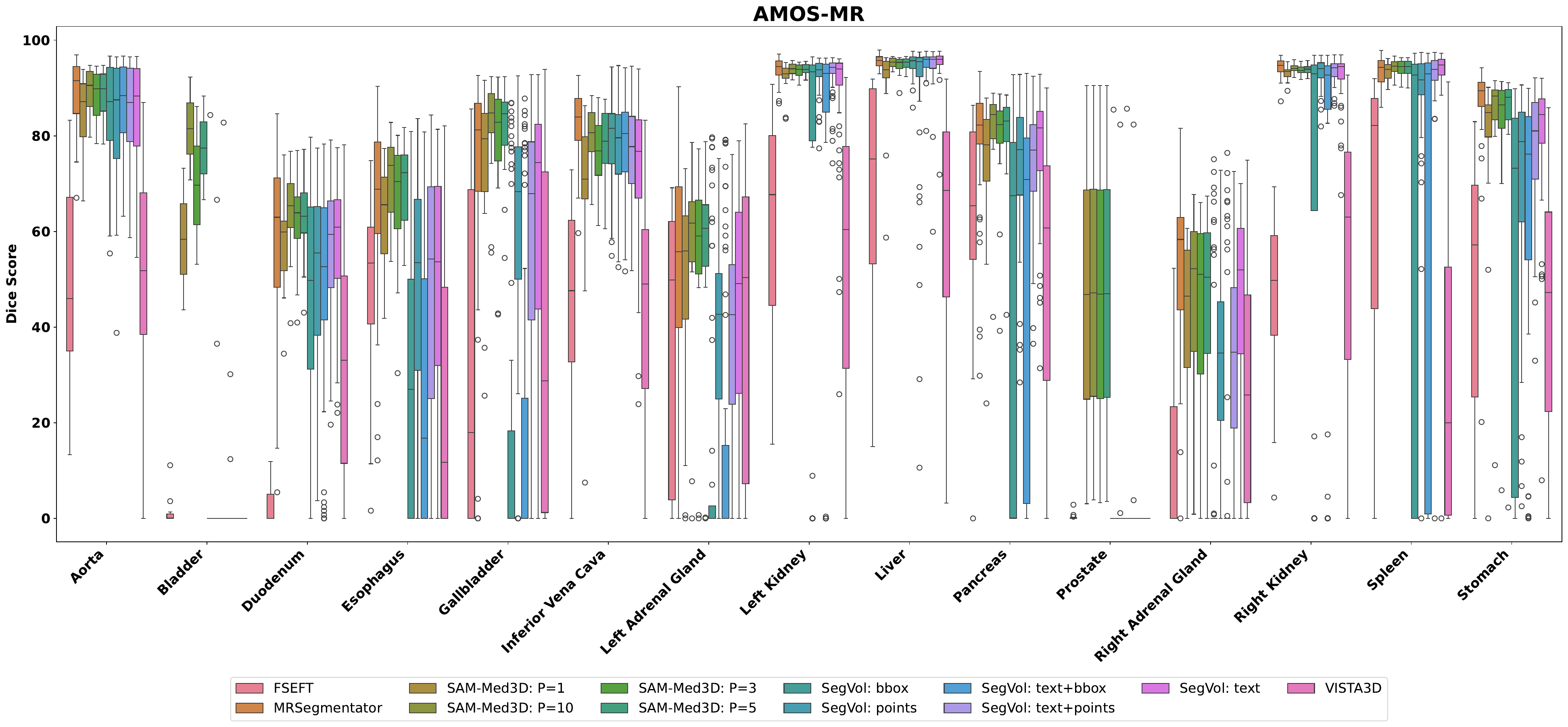}
    }
    \makebox[\textwidth][c]{%
        \includegraphics[width=1.2\textwidth]{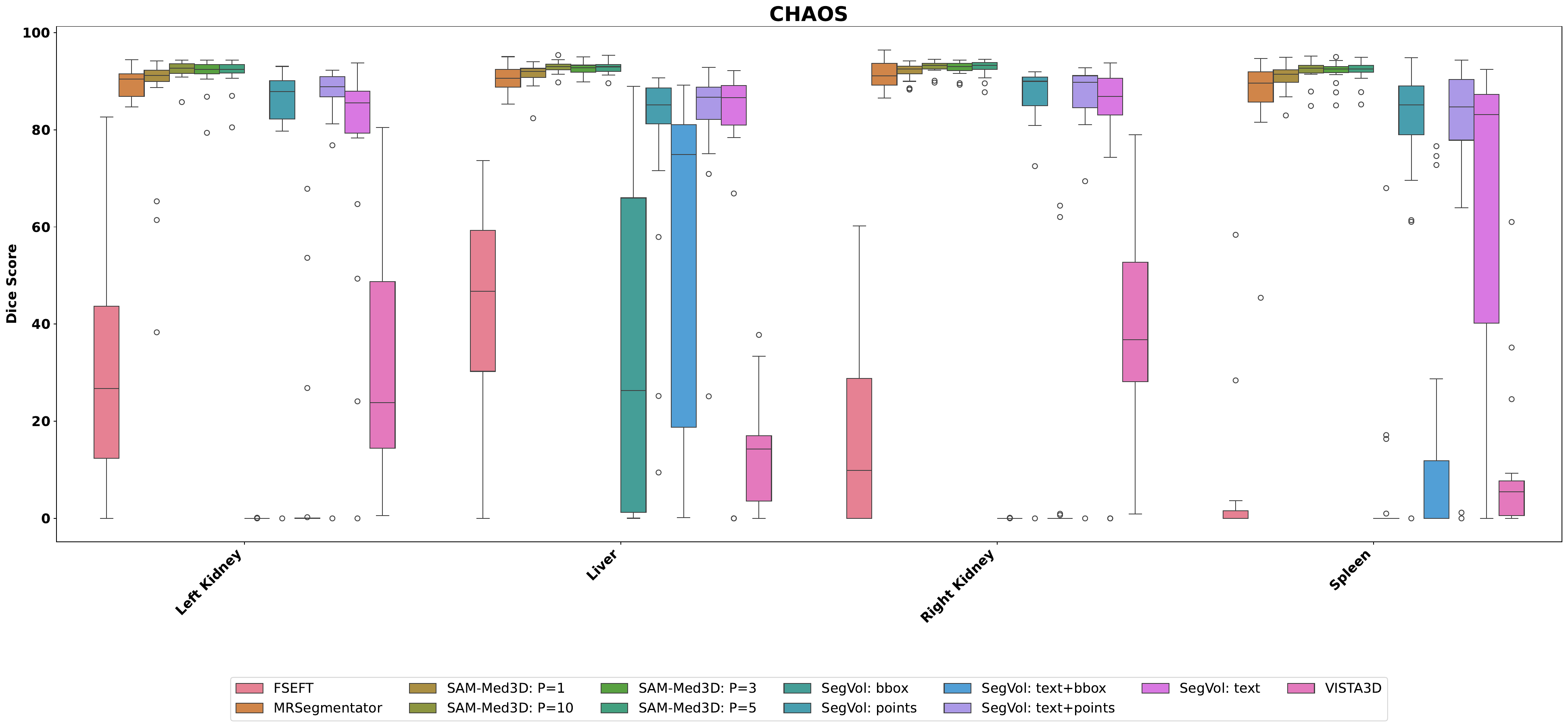}
    }
    \caption{Box plots for organ-wise dice scores for all comparative models on the \textbf{abdominal MR} datasets used in this study.}
    \label{fig:organwise_abdmr}
\end{figure}

\begin{figure}
    \centering
    \makebox[\textwidth][c]{%
        \includegraphics[width=1.2\textwidth]{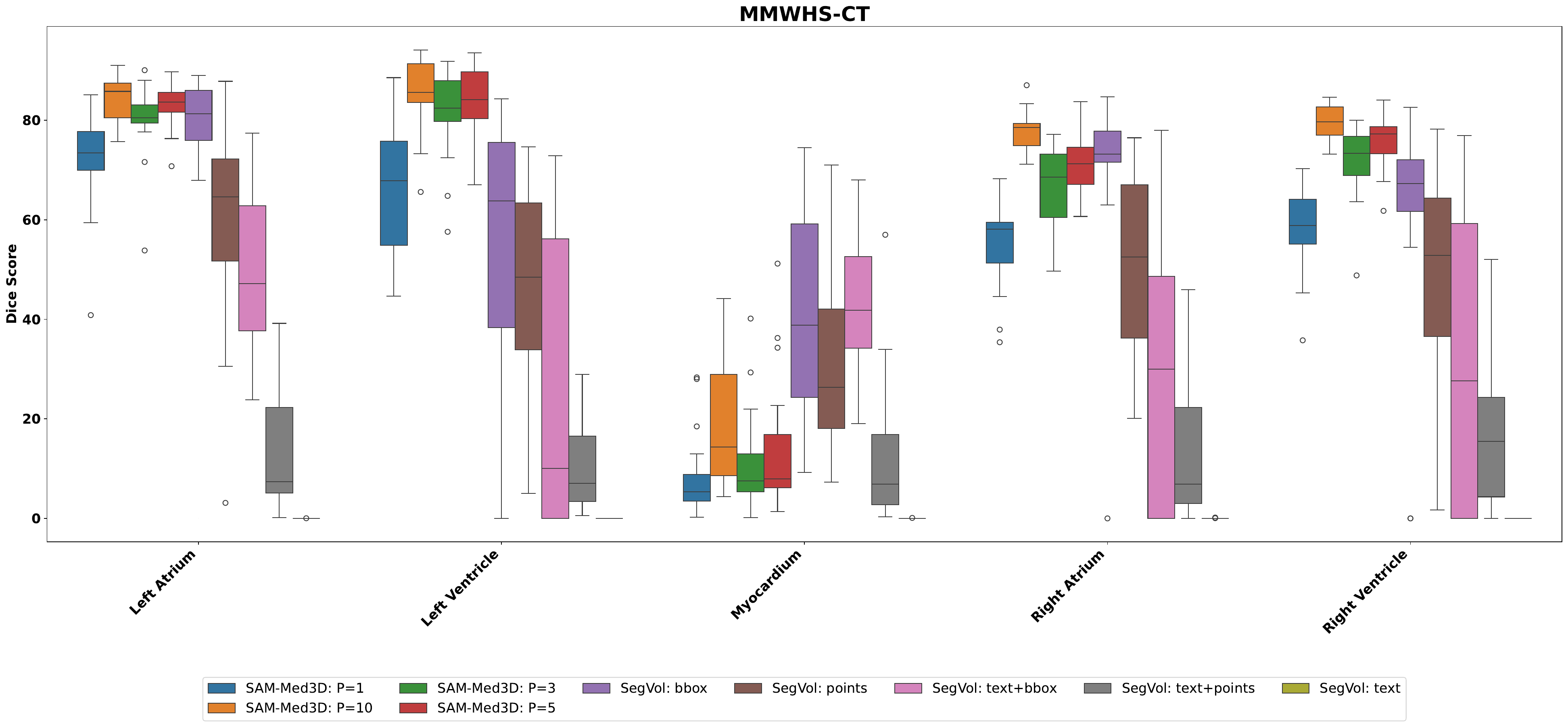}
    }
    \makebox[\textwidth][c]{%
        \includegraphics[width=1.2\textwidth]{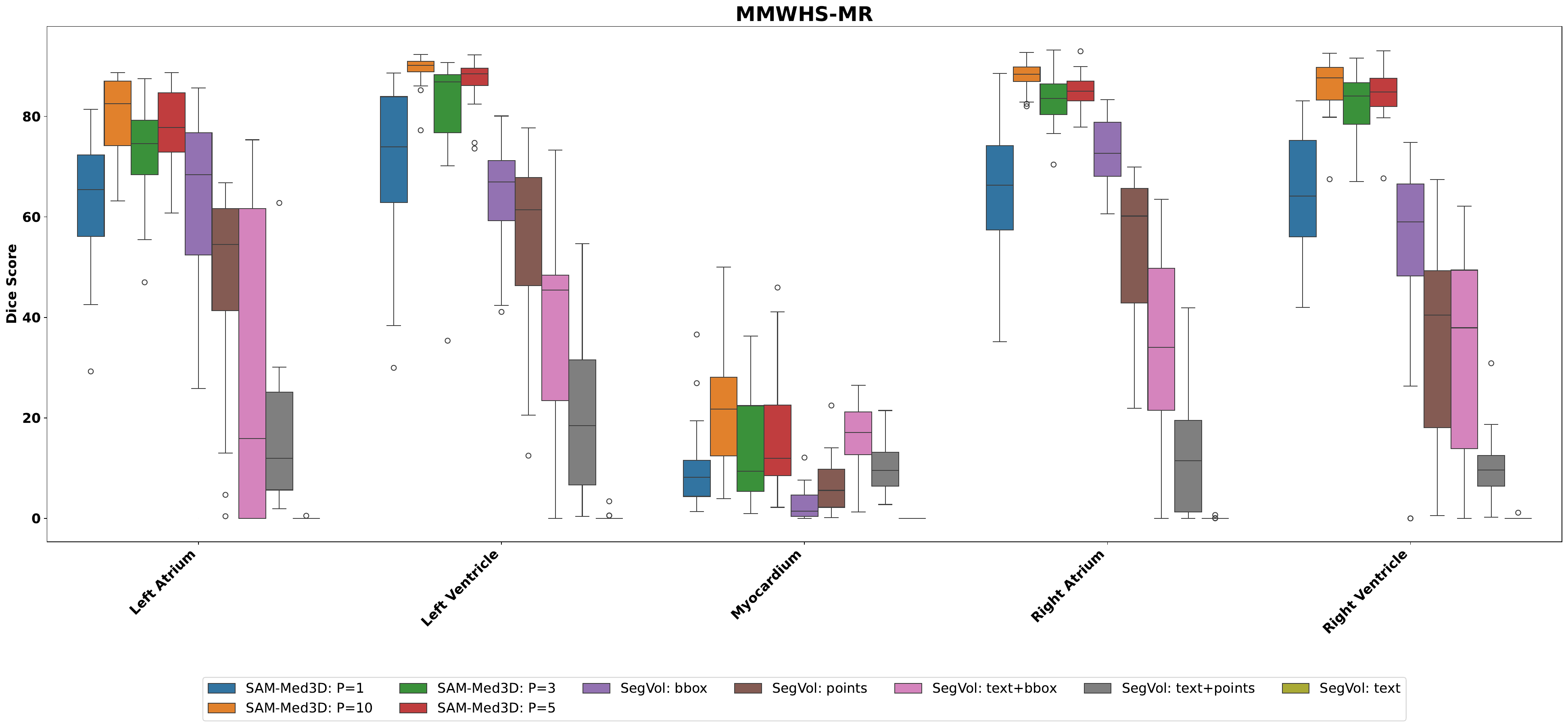}
    }
    \caption{Box plots for organ-wise dice scores for all comparative models on the \textbf{cardiac} datasets used in this study.}
    \label{fig:organwise_cardiac}
\end{figure}

For additional analysis, we also provide organ-wise segmentation results of the comparative models across the \textit{multi-class} datasets used in this study. We show these results using box plots in \autoref{fig:organwise_abdct} for abdominal CT, \autoref{fig:organwise_abdmr} for abdominal MR, and \autoref{fig:organwise_cardiac} for the cardiac datasets. 

From the figures, we observe that for certain organs like liver, kidneys and spleen, the segmentation performance across foundational models is more reliable (both in-domain and out-of-domain) compared to other structures (duodenum or adrenal glands). Notably, the training set of models like FSEFT~\cite{fseft} or SegVol~\cite{segvol} also comprised several organ-specific datasets, such as the Medical Segmentation Decathlon datasets with liver, pancreas and spleen segmentation tasks~\cite{decathlon}, Liver Tumor Segmentation (LiTS'17)~\cite{lits}, Kidney Tumor Segmentation (KiTS'19, KiTS'23) \cite{kits19,kits23}, among others. We hypothesize that the superior performance of models on these anatomical structures is due to their higher frequency of occurrence in their respective training corpora.

\end{document}